\documentclass[conference]{IEEEtran}
\ifCLASSINFOpdf
\else
\fi
\usepackage{microtype}
\usepackage{graphicx}
\usepackage{color,hyperref}
\usepackage{amsmath,amssymb}
\usepackage{amsmath}
\usepackage{lipsum}
\usepackage{algorithmicx}

\usepackage{subcaption}
\usepackage{comment}
\usepackage{url}
\usepackage{multirow}
\usepackage{adjustbox}                          
\usepackage{adjustbox}
\DeclareUnicodeCharacter{2212}{-}

\begin{document}
%
\title{\textit{Do You Do Yoga?} Understanding Twitter Users' Types and Motivations using Social and Textual Information}

\author{\IEEEauthorblockN{Tunazzina Islam}
\IEEEauthorblockA{\textit{Department of Computer Science} \\
\textit{Purdue University}\\
West Lafayette, IN, USA \\
islam32@purdue.edu}
\and
\IEEEauthorblockN{Dan Goldwasser}
\IEEEauthorblockA{\textit{Department of Computer Science} \\
\textit{Purdue University}\\
West Lafayette, IN, USA  \\
dgoldwas@purdue.edu}
}
\maketitle

\begin{abstract}
Leveraging social media data to understand people's lifestyle choices is an exciting domain to explore but requires a multiview formulation of the data. In this paper, we propose a joint embedding model based on the fusion of neural networks with attention mechanism by incorporating social and textual information of users to understand their activities and motivations. We use well-being related tweets from Twitter, focusing on \textit{`Yoga'}. We demonstrate our model on two downstream tasks: (i) finding user type such as either practitioner or promotional (promoting yoga studio/gym), other; (ii) finding user motivation i.e. health benefit, spirituality, love to tweet/retweet about yoga but do not practice yoga.
\end{abstract}
\section{Introduction}
In recent years social media platforms emerged as the primary space for people to interact with their friends and share ideas. Platforms, such as Twitter, represent a largely untapped resource for understanding lifestyle, well-being, and health choices \cite{prieto2014twitter,amir2017quantifying,schwartz2016predicting,yang2016life,islam2019yoga}. In this paper, we look at `yoga', as an example of a popular and multi-faceted activity. Many studies show that yoga can alleviate symptoms of anxiety and depression as well as promote good physical fitness \cite{khalsa2004treatment,ross2010health}. Interest in this topic can come from practitioners or commercial parties. Furthermore, practitioners can be motivated by different reasons like staying healthy physically and mentally or spiritual practice. 

The main motivation of this paper is to understand lifestyle choices and model their connection to other indicators, such as mental health, decision making etc. This type of analysis depends on capturing relevant information about yoga activity from the text at the appropriate level of granularity--  understanding the outcomes of lifestyle choices requires modeling the people's motivation for engaging in these activities, their level of commitment. Unfortunately, the short and often ambiguous nature of tweets makes our analysis challenging. As a result, simple pattern-based analysis using yoga-related keywords often falls short of capturing this information. Consider these two examples:
\newline
{\small \textbf{Tweet 1:} \textit{Scorpion pose or Vrischikasana with a wall. Love this feeling. \#yoga \#scorpionpose}} 
\newline
{\small \textbf{Tweet 2:} \textit{Gentle \#Yoga \& Meditation w/ Michael. BreakOut Studios Online Classes.}}

The same ``\#yoga” is used by two different types of users. The first tweet is about doing a specific yoga pose and the user's feelings about it. The second tweet is about online yoga classes in studios. Our main insight in this paper is that understanding user types and motivation should be done collectively over both the tweets they author, their profile information, and social behavior. For example, the profile description of the user of Tweet 1 is {\small \textit{``@UCLA alumni, traveler, yoga and nature lover."}} that indicates the user as a practitioner. On the other hand, the profile description of the user of Tweet 2 is {\small \textit{``A welcoming dance \& fitness facility for movers ages 13+. Streaming classes daily on YouTube"}} which is an indicator of being a promotional account.

Moreover, the users' social behavior indicated via their social network can help further disambiguate the tweet text. Based on the principle of homophily \cite{mcpherson2001birds}, the users' activity type and motivation for it, is likely to be reflected by their social circles, often having similar attributes. Our main technical contribution is to suggest a model for aggregating users' tweets and contextualizing this textual content with social information.

While there has been a significant amount of work dedicated to understanding the demographic properties of Twitter users from a combination of their tweets and social information \cite{li2015learning,benton2016learning,mishra2018neural,islam2020does}, our main challenge is to use this information to construct a coherent user representation relevant for the specific set of choices we are interested in. We suggest a method for combining a large amount of Twitter content and social information associated with each user by building a joint embedding attention-based neural network model. We refer to our model as Yoga User Network (YUN)\footnote[1]{Code and data at \url{https://github.com/tunazislam/Yoga-User-Network-YUN}}. Using this model, we predict user type who tweets about yoga whether they are practitioner, promotional, or others and find user motivation such as some people do yoga or tweet about yoga for health benefit, some do it for a spiritual retreat, others i.e. appreciate yoga but do not practice. We use accuracy and macro average F1-score as the evaluation metrics of our model. We evaluate YUN under ten different baselines: {\small(i) Description; (ii) Location; (iii) Tweets; (iv) Network; (v) BERT (Bidirectional
Encoder Representations from Transformers) \cite{devlin2018bert} fine-tuned with Description (Description\_BERT); (vi) BERT fine-tuned with Location (Location\_BERT); (vii) BERT fine-tuned with Tweets (Tweets\_BERT); (viii) joint embedding on description and location (Des + Loc); (ix) joint embedding on description, location, and tweets (Des + Loc + Twt); (x) joint embedding on description, location, and network (Des + Loc + Net).} We show that YUN (Des + Loc + Twt + Net) outperforms these baselines.

\begin{figure}[htbp]
  \centering  
  \includegraphics[width= 0.5 \textwidth]{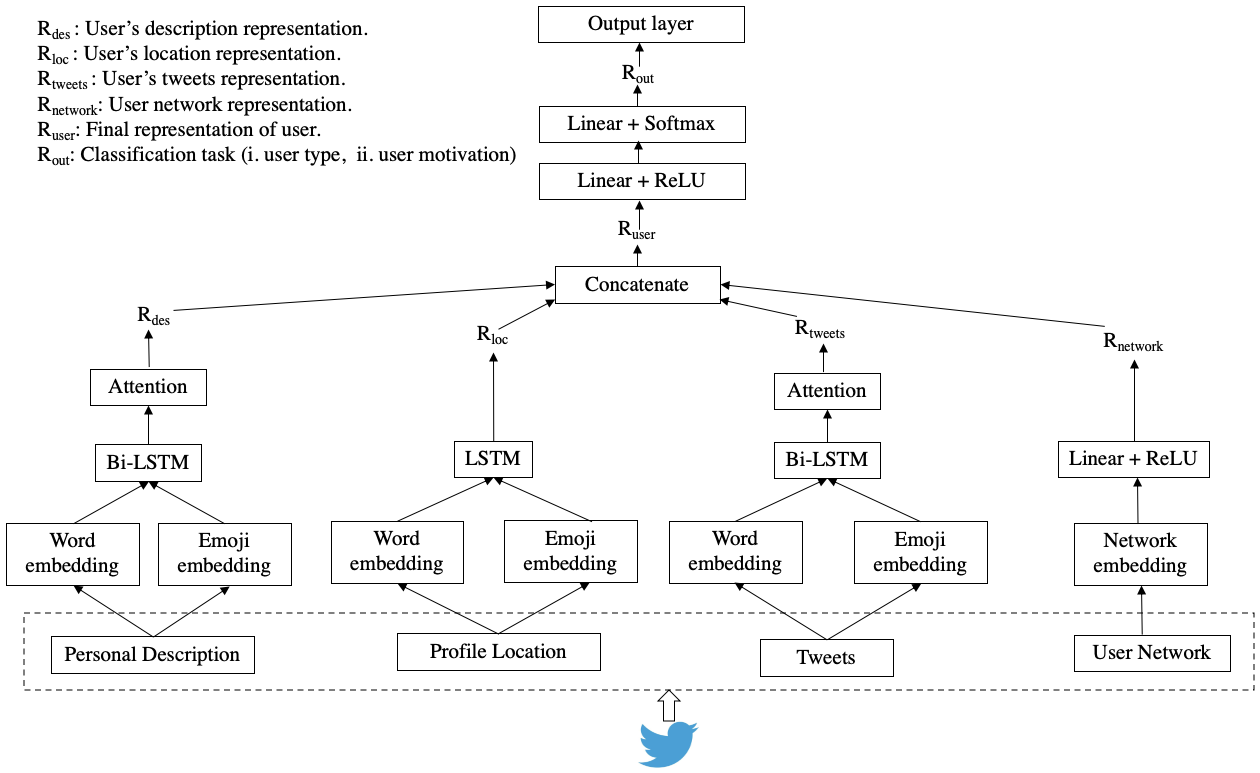}
    \caption{Overview of the YUN model. YUN has four sub-networks: Description, Location, Tweets, and User Network.}
    \label{fig:model}
\end{figure}
\vspace{-11pt}
\section{Technical Approach}
Our goal is to learn user representation by leveraging tweet text, emoji, metadata, and user network. Then we use this model for the downstream classification task. Fig. \ref{fig:model} shows the overall architecture of our proposed model. 
\subsection{Tweet Representation}
\vspace{-2pt}
Tweets contain text or/and emoji. Suppose that $T = \{w_1, w_2, ..., w_T\}$ denotes a tweet generated by a user $u$. To represent the tweet's textual content, we use pre-trained word embeddings Word2Vec ($300d$) using Skip-gram architecture \cite{mikolov2013efficient}. For emoji embeddings, we use $300d$ Emoji2Vec \cite{eisner2016emoji2vec}. The embeddings of text and emoji are then forwarded to a Bi-LSTM \cite{hochreiter1997long}. It produces the hidden representation $H = (h_1, h_2, ...,h_n)$, where $h_i$ is the hidden
state of the LSTM at time-step $i$, summarizing all
the information of the tweet up to $w_i$. We get the final hidden representation by concatenating the both directions, 
\vspace{-5pt}
\begin{align}
h_i = \overrightarrow{h_i} || \overleftarrow{h_i}, h_i \in \mathbb{R}^{2D}\label{eq:5},
\end{align}
\vspace{-20pt}
\newline
where, $\rightarrow$ denotes forward, $\leftarrow$ is backward and $D$ is the size of each LSTM.

To assign important words in the final representation, we use a context-aware attention mechanism \cite{bahdanau2014neural} which assigns a weight, $a_i$ to each word. We compute the representation of the tweets, $R_{tweets}$ which is the weighted sum of all word and emoji representations using attention weights and defined as
\vspace{-6pt}
\begin{align}
R_{tweets} &= \sum_{i=1}^{T} a_ih_i, R_{tweets} \in \mathbb{R}^{2D}\label{eq:6},
\end{align}
where 
\vspace{-17pt}
\begin{align}
a_i &= \frac{exp(m_ic_h)}{\sum_{t=1}^T exp(m_tc_h)}\label{eq:7}, \\
m_i &= tanh(W_hh_i + b_h), m_i \in [-1, 1] \label{eq:8},
\end{align}
\vspace{-20pt}
\newline
and $W_h$ and $c_h$ are the layer's weights and $b_h$ is the bias.
\subsection{Metadata Representation}
\vspace{-2pt}
In this work, we consider two metadata fields i.e. user description and location. For description, embeddings of text and emoji are forwarded to Bi-LSTM attention layer to build the final representation of user description, $R_{des} \in \mathbb{R}^{2D}$ by implementing similar approach followed in $R_{tweets}$.

As the length of sequence for location is shorter, embeddings of text and emoji in location are fed to a LSTM to obtain the representation of user location, $R_{loc}\in \mathbb{R}^{D}$, where $D$ is the size of LSTM.
\subsection{User Network Representation}
\vspace{-2pt}
User network is constructed based on $@-$mentions in users' tweets \cite{rahimi2015exploiting}. We create a graph from interactions among users via retweets and/or $@-$mentions. In this graph, nodes are all users in our dataset, as well as other external users mentioned in users' yoga-related tweets. An undirected and unweighted edge is created between two users if either user mentions the other. In this work, we do not consider edge weights and self-loop. We have $36642$ nodes and $50729$ edges.

We model user network using graph, $G = (V, E)$ where $V$ is the set of nodes representing users and $E$ is the set of edges representing interactions among users. Here, $u_i \in V$ refers to a node in $G$, and $e_{ij} \in E$ denotes the edge connecting nodes $u_i$ and $u_j$. To compute node embedding, we use Node2Vec \cite{grover2016node2vec}. Similar to word2vec's Skip-gram model, for every node $u$, Node2Vec creates a mapping function, $f : V \longrightarrow \mathbb{R}^d$ which maps $u$ to a low dimensional embedding of size $d$. Here, $f$ is a matrix of size $|V| \times d$ parameters. For every source node, $u \in V$, there is a network neighborhood, $N_S(u) \subset V$ of node $u$ generated through a neighborhood sampling strategy, $S$. The mapping function, $f$ maximizes the probability of observing nodes belonging to $S(u)$ which is the set of $n$ nodes encountered in the graph by taking $k$ random walks starting from $u$. Thus, we generate the embedding of user network, $E_{net} = (e_{u_1}, e_{u_2},..., e_{u_V})$ and forward to linear layer with \textit{ReLU} \cite{nair2010rectified} activation function to compute user network representation, $R_{net} \in \mathbb{R}^d $ which is defined as:
\vspace{-5pt}
\begin{align}
R_{net} = relu(W_{n} E_{net} + b_{n}), \label{eq:9}
\end{align}
\vspace{-20pt}
\newline
where, $W_n$ and $b_n$ are the layer's weight and bias.
\subsection{User Representation}
\vspace{-2pt}
Concatenating the four representations generated from the four sub-networks (Fig. \ref{fig:model}), we create the final user representation, $R_{user}$. We define $R_{user} \in \mathbb{R}^{(2D+D+2D+d) }$ as follows:
\vspace{-15pt}
\begin{align}
R_{user} = R_{des} || R_{loc} || R_{tweets} || R_{net} \label{eq:10}
\end{align}
\vspace{-20pt}
\newline
We denote the concatenation operation as $||$. We pass $R_{user}$ through a fully connected two-layer classifier.
\vspace{-20pt}
\section{Experimental Settings}
\vspace{-2pt}
\subsection{Dataset Details}
\vspace{-2pt}
We use Twitter streaming API to extract 419608 tweets sub-sequentially from May to November, 2019 related to yoga containing specific keywords. We randomly select $1300$ users having $3097678$ timeline tweets.
\begin{table}[htbp]
  \centering
  \caption{A summary of hyperparameter settings of all models}
  \begin{adjustbox}{width=\columnwidth,center}
    \begin{tabular}{|c|c|c|c|c|c|c|c|c|c|c|}
    \hline
    \textbf{Model} & \textbf{lr}  & \textbf{opt} & \textbf{reg} & \textbf{word}  & \textbf{emoji}   & \textbf{lstm}  & \textbf{attn} & \textbf{h} & \textbf{eut} & \textbf{eum}\\
    \hline
    Description  & 0.01 & Adadelta & .0001 & 300 & 300 & 150 & 300 & 200 & 22 & 14\\
    \hline
    Location & 0.01 & SGD momentum & N/A & 300 & 300 & 150  & 300 & 200 & 8 & 7\\
    \hline
    Tweets  & 0.01 & Adadelta & .0001 & 300 & 300 & 150 & 300 & 200 & 11 & 11 \\
    \hline
     Network  & 0.01 & Adadelta  & .0001 & N/A & N/A & N/A & N/A & 200 & 23 & 23\\
    \hline
    Description\_BERT  & 0.00002 & AdamW & .01 & N/A & N/A & N/A & N/A & N/A & 3 & 3  \\
    \hline
    Location\_BERT  & 0.00002 & AdamW & .01 & N/A & N/A & N/A & N/A & N/A & 1 & 1  \\
    \hline
    Tweets\_BERT  & 0.00002 & AdamW & .01 & N/A & N/A & N/A & N/A & N/A & 1 & 3  \\
    \hline
    Des + Loc  & 0.01 & Adadelta & .0001 & 300 & 300 & 150 & 300 & 200 & 13 & 21\\
    \hline
    Des + Loc + Twt  & 0.01 & Adadelta & .0001 & 300 & 300 & 150 & 300 & 200 & 16 & 25 \\
    \hline
    Des + Loc + Net  & 0.01 & Adadelta & .0001 & 300 & 300 & 150 & 300 & 200 & 15 & 13 \\
    \hline
    \textbf{YUN}  & 0.01 & Adadelta & .0001 & 300 & 300 & 150 & 300 & 200 & 17 & 18 \\
    \hline
    \hline
    \multicolumn{11}{|p{40em}|}{lr : Learning rate.} \\
    \multicolumn{11}{|p{40em}|}{opt: Optimizer.} \\
    \multicolumn{11}{|p{40em}|}{reg : Weight decay ($L^2$ regularization).} \\
    \multicolumn{11}{|p{40em}|}{word: Word embedding dimension.} \\
    \multicolumn{11}{|p{40em}|}{emoji: Emoji embedding dimension.} \\
    \multicolumn{11}{|p{40em}|}{lstm: LSTM unit size.} \\
    \multicolumn{11}{|p{40em}|}{attn: Attention vector size.} \\
    \multicolumn{11}{|p{40em}|}{h: Size of $L_{user}$ layer which is the first layer of two-layer classifier. } \\
    \multicolumn{11}{|p{40em}|}{eut: Best result achieved at epochs for user type classification. Epoch starts with $0$.} \\
    \multicolumn{11}{|p{40em}|}{eum: Best result achieved at epochs for user motivation classification. Epoch starts with $0$.} \\
     \hline
    \end{tabular}
    \end{adjustbox}
  \label{tab:hyperparam}
\end{table}
For pre-processing the tweets, we first convert them into lower case, remove URLs, keep emojis, then use a tweet-specific tokenizer from NLTK to tokenize them. After removing stop words, we use WordNetLemmatizer of NLTK for lemmatization. We manually annotate $1300$ users based on the intent of the tweets and observation of user description. Consider the tweet {\small \textit{``Yoga is more than fitness, it’s a mental and spiritual release. \#MyMondayMotivation \#Yoga"}}. We annotate this tweet user as a \textit{practitioner} having \textit{spiritual} motivation.
In our user type annotated data, we have $42\%$ practitioner, $21\%$ promotional, and $37\%$ other users. For user motivation, we have $51\%$ users who tweet about yoga regarding health benefit, $5\%$ spiritual, and $41\%$ other motivation. We shuffle the dataset and then split it into train ($60\%$), validation ($20\%$) and test ($20\%$).
\subsection{Hyperparameter Details}
\vspace{-2pt}
We perform grid hyperparameter search on validation set for all models except the BERT fine-tuned models. We explore $0.005, 0.01, 0.05, 0.1, 0.5$ learning rate values and $0, 10^{-4}, 10^{-3}, 10^{-2}$ for weight decay. We run the models total $30$ epochs and plot curves for loss and macro-avg F1 score. Our early stopping criteria is based on steeply increased validation loss point.
For Description\_BERT, Location\_BERT, and Tweets\_BERT model, we use pre-trained and fine-tuned \textit{BertForSequenceClassification} model. To encode texts, we choose maximum sentence length = $160, 50, 500$ respectively for padding or truncating, batch size = $32$, learning rate = $2e-5$, optimizer = AdamW \cite{loshchilov2017decoupled}, epsilon parameter = $1e-8$, number of epochs = $4$.
For network embedding, we use Node2Vec with following parameters: dimension = $300$, number of walks per source = $10$, length of walk per source = $80$, window size = $10$, min count = $1$. We put network embedding vectors as $0$ for not appearing users in the mention network. We pass the network embedding to a linear layer of size $150$ activated by \textit{ReLU}. We set the embedding vectors $0$ if users do not have location and/or description.
In YUN model, final user representation is passed through a linear layer with dimension $600$ and activated by \textit{ReLU} and the output of this layer is fed to another linear layer with dimension $200$ and activated by \textit{softmax}. We set the hyper-parameters of YUN model as follows:
learning rate = $0.01$, epochs = $17$ (user type); epochs = $18$ (user motivation), optimizer = Adadelta \cite{zeiler2012adadelta}, weight decay = $10^{-4}$. Table \ref{tab:hyperparam} provides a brief summary of hyperparameter settings of all models.
\begin{figure}
\begin{subfigure}{.5\columnwidth}
  \centering
  \includegraphics[width=1\textwidth]{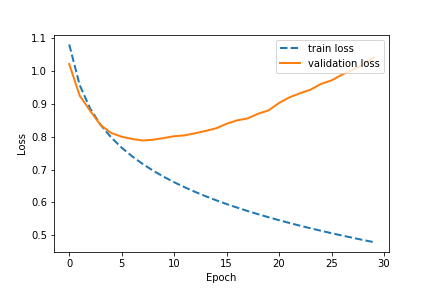}
  \caption{}
  \label{fig:yun_ut_loss}
\end{subfigure}%
\begin{subfigure}{.5\columnwidth}
  \centering
  \includegraphics[width=1\textwidth]{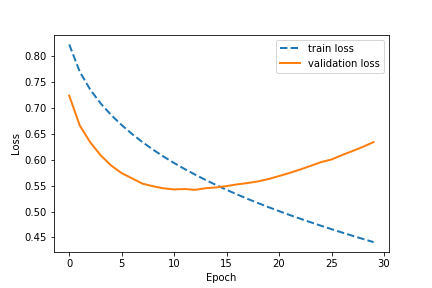}
  \caption{}
  \label{fig:yun_um_loss}
\end{subfigure}
\begin{subfigure}{.5\columnwidth}
  \centering
  \includegraphics[width=1\textwidth]{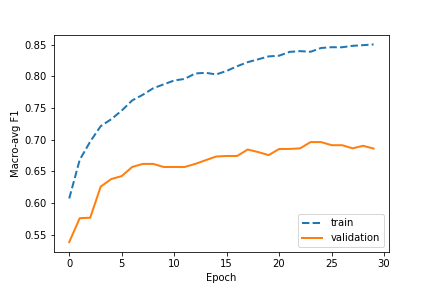}
  \caption{}
  \label{fig:yun_ut_f1}
\end{subfigure}%
\begin{subfigure}{.5\columnwidth}
  \centering
  \includegraphics[width=1\textwidth]{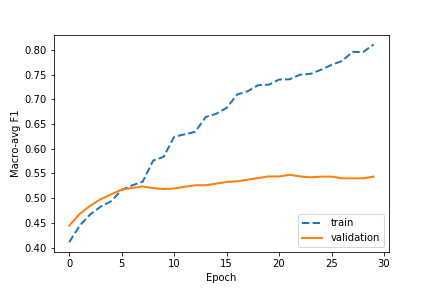}
  \caption{}
  \label{fig:yun_um_f1}
\end{subfigure}
\caption{Learning curves of YUN model for training and validation data. (a) loss vs. epochs for user type, (b) loss vs. epochs for user motivation, (c) macro-avg F1 score vs. epochs for user type, (d) macro-avg F1 score vs. epochs for user motivation.}
\label{fig:yun_loss_f1}
\end{figure}
\begin{table}
\caption{Performance comparisons on test data}
\begin{adjustbox}{width=\columnwidth,center}
\begin{tabular}{|c|c|c||c|c|}
          \cline{2-5} \multicolumn{1}{c|}{} & \multicolumn{2}{c||}{\textbf{user type}} & \multicolumn{2}{c|}{\textbf{user motivation}} \\
          \hline
    \textbf{Model} & \textbf{Accuracy}  & \textbf{Macro F1 score}  & \textbf{Accuracy}  & \textbf{Macro F1 score}  \\
    \hline
    Description  & 0.725 & 0.693 & 0.782 & 0.575 \\
    \hline
    Location  & 0.676 & 0.563 & 0.695 & 0.470 \\
    \hline
    Tweets  & 0.721 & 0.687  & 0.744 & 0.504 \\
    \hline
    Network  & 0.752 & 0.557  & 0.790 & 0.541 \\
    \hline
    Description\_BERT  &  0.718 & 0.681 &  0.771 & 0.528 \\
    \hline
    Location\_BERT  &   0.679 & 0.606 & 0.695 & 0.476 \\
    \hline
    Tweets\_BERT  & 0.760 &  0.669  &  0.805 &  0.551 \\
    \hline
    Des + Loc & 0.733 & 0.693 & 0.775 & 0.599 \\
    \hline
    Des + Loc + Twt  & 0.760 & 0.725 & 0.794 & 0.580 \\
    \hline
    Des + Loc + Net & 0.775 & 0.723 & 0.828  & \textbf{0.647}  \\
    \hline
    \textbf{YUN} & \textbf{0.790} & \textbf{0.742} & \textbf{0.844}  & 0.619  \\
    \hline
    \end{tabular}
\end{adjustbox}
\label{tab:result}
\end{table}
\subsection{Results}
\vspace{-2pt}
We evaluate our method using two commonly used metrics: Accuracy and Macro-average F1-score. Table \ref{tab:result} shows the performance comparison of YUN model with ten baselines on test dataset. YUN achieves the highest test accuracy $(79.0\%)$ and macro-avg F1 score $(74.2\%)$ for classifying user type. We achieve comparable performance for classifying user motivation where YUN obtains the highest test accuracy $(84.4\%)$  but Des + Loc + Net model achieves better macro-avg F1 score $(64.7\%)$ than YUN $(61.9\%)$. Fig. \ref{fig:yun_loss_f1} shows the learning curves loss (train and validation) vs. epochs and macro-avg F1 score (train and validation) vs. epochs for YUN model.
\subsection{Ablation Study}
\vspace{-2pt}
In this section, we provide an ablation study to evaluate the contribution of each model component in classifying user type and motivation. We train an individual neural network model for each field i.e. description, location, tweets and network. At the end, we feed the final representation of each sub-network to a two layer classifier activated by \textit{ReLU} and then \textit{softmax} function. The $1^{st}$, $2^{nd}$, $3^{rd}$, $4^{th}$ rows of the Table \ref{tab:result} show the performance breakdown for each model over the test dataset.
The results conclude that profile description of user is the most informative field for understanding user type and motivation, and model trained on this source achieves the best single source performance based on macro-avg F1 score (user type: $69.3\%$ and user motivation: $57.5\%$). This model can correctly predict $72.5\%$ user type and $78.2\%$ user motivation (Table \ref{tab:result}).

User network model can correctly classify $75.2\%$ user type and $79.0\%$ user motivation. However, our experiments
show that excluding user network information (Des + Loc + Twt model) declines the performance of the final model in terms of both accuracy by $3.8\%$ for user type \& $5.9\%$ for user motivation and macro-avg F1 score by $2.3\%$ for user type \& $6.3\%$ for user motivation (Table \ref{tab:result}). 
Model using only metadata fields (Des + Loc) provides an accuracy $73.3\%$ for user type and $77.5\%$ for user motivation where the description field is the most informative one (Table \ref{tab:result}). User location, on the other hand, has the minimum accuracy in predicting user type and motivation. 
\subsection{Error Analysis}
\vspace{-2pt}
As reported in the Table \ref{tab:result}, our proposed YUN model achieves the best test accuracy and macro-avg F1 score for classifying user type. But for user motivation classification task, YUN has less macro-avg F1 score than Des + Loc + Net model. Adding tweets information decreases the macro-avg F1 score as our tweets representation does not have contextualized word embedding. Imbalanced data distribution for user motivation could be another reason for that as only $5\%$ users of our data has spiritual motivation. 

Our ablation study demonstrates that the profile description field highly contributes to the classification task for user types and motivations. However, some prediction errors arise when description fields are absent or misleading. We found two cases that result misclassification: (1) Users do not have their profile description in Twitter; (2) Users who provide profile description related to ``yoga" but they usually just retweet yoga-related quotes. From our ablation study, we notice that the user location has relatively low accuracy and macro-avg F1 score. One of our assumptions is to understand user type and motivation based on geographic location. Users coming from the same geographic location might have a similar motivation to follow a certain lifestyle choice. We consider location embeddings but users sometimes do not provide location information on Twitter.
\section{Conclusion}
\vspace{-2pt}
We suggest a joint embedding attention-based neural network model called YUN, which explicitly learns Twitter user representations by leveraging social and textual information of users to understand their types and motivations. Our model can be used to understand users’ type and motivation for different lifestyle choices i.e. ``keto diet”, ``veganism”. In future, we aim to develop a contextualized model to predict user type and motivation using minimal supervision.
\bibliography{yun.bib}{}
\bibliographystyle{ieeetr}
\end{document}